\documentclass[final]{cvpr}

\usepackage{times}
\usepackage{epsfig}
\usepackage{graphicx}
\usepackage{amsmath}
\usepackage{amssymb}
\makeatletter
\@namedef{ver@everyshi.sty}{}
\makeatother
\usepackage{tcolorbox}
\usepackage{booktabs}

\usepackage[pagebackref=true,breaklinks=true,colorlinks,bookmarks=false]{hyperref}

\def\cvprPaperID{13} 
\def\confYear{CVPR 2021}

\begin{document}

\title{Neural Architecture Search of Deep Priors: \\ Towards Continual Learning without Catastrophic Interference}

\author{Martin Mundt$^{*}$, Iuliia Pliushch\thanks{equal contribution}\, , Visvanathan Ramesh\\
Goethe University, Frankfurt, Germany\\
{\tt\small \{mmundt, pliushch, vramesh\}@em.uni-frankfurt.de}
}

\maketitle

\begin{abstract}
In this paper we analyze the classification performance of neural network structures without parametric inference. Making use of neural architecture search, we empirically demonstrate that it is possible to find random weight architectures, a deep prior, that enables a linear classification to perform on par with fully trained deep counterparts. Through ablation experiments, we exclude the possibility of winning a weight initialization lottery and confirm that suitable deep priors do not require additional inference. In an extension to continual learning, we investigate the possibility of catastrophic interference free incremental learning. Under the assumption of classes originating from the same data distribution, a deep prior found on only a subset of classes is shown to allow discrimination of further classes through training of a simple linear classifier.
\end{abstract}

\section{Introduction}
Prevalent research routinely inspects continual deep learning through the lens of parameter inference. As such, an essential desideratum is to overcome the threat of \textit{catastrophic interference} \cite{McCloskey1989, Ratcliff1990}. The latter describes the challenge to avoid accumulated knowledge  from being continuously overwritten through updates on currently observed data instances. As outlined by recent reviews in this context \cite{Parisi2019, Mundt2020}, specific mechanisms have primarily been proposed in incremental classification scenarios. Although precise techniques vary drastically across the literature, the common focus is a shared goal to maintain a deep encoder's representations, in an effort to protect performance from continuous degradation \cite{Lee2017, Kirkpatrick2017, Chaudhry2018, Rebuffi2017, Nguyen2018, Shin2017, Ostapenko2019, Mundt2019, Ayub2021}. 

In this work, we embark on an alternate path, one that asks a fundamentally different question: \textit{What if we didn't need to infer a deep encoder's parameters and thus didn't have to worry about catastrophic interference altogether?}
This may initially strike the reader as implausible. However, successfully solving the discriminative classification task has but one intuitive requirement: data belonging to different classes must be easily separable. \\
Inspired by the possibility of recovering encodings generated by random projections in signal theory \cite{Candes2006}, selected prior works have thus investigated the utility of (neural) architectures with entirely random weights \cite{Saxe2011, Huang2011}, see \cite{Cao2018} for a recent review. Their main premise is that if a deep network with random weights is sensitive to the low-level statistics, it can progressively enhance separability of the data and a simple subsequent linear classification can suffice \cite{Giryes2016}. Whereas promising initial results have been presented, a practical gap to fully trained systems seems to remain. We posit that this is primarily a consequence of the complexity involved in hierarchical architecture assembly.

In the spirit of these prior works on random projections, we frame the task of finding an adequate deep network with random weights, hence referred to as a \textit{deep prior}, from a perspective of neural architecture search (NAS) \cite{Baker2016, Real2017, Zoph2017}. We empirically demonstrate that there exist configurations that achieve rivalling accuracies to those of their fully trained counterparts. We then showcase the potential of deep priors for continual learning. We structure the remainder of the paper according to our  contributions: \\
1. We formulate deep prior neural architecture search (DP-NAS) based on random weight projections. We empirically show that it is possible to find hierarchies of operations which enable a simple linear classification. As DP-NAS does not require to infer the encoders' parameters, it is magnitudes of order faster than conventional NAS methods. \\
2. Through ablation experiments, we observe that deep priors are not subject to a weight initialization lottery. That is, performance is consistent across several randomly drawn sets of weights. We then empirically demonstrate that the best deep priors capture the task through their structure. In particular, they do not profit from additional training.  \\
3. We corroborate the use of deep priors in continual learning. Without parameter inference in the deep architecture, disjoint task settings with separate classifiers per task are trivially solved by definition. In scenarios with shared classifier output, we empirically demonstrate that catastrophic interference can easily be limited in a single prediction layer. Performances on incremental MNIST \cite{LeCun1998}, FashionMNIST \cite{Xiao2017} and CIFAR-10 \cite{Krizhevsky2009} are shown to compete with complex deep continual learning literature approaches. 

We conclude with limitations and prospects.

\section{Deep Prior Neural Architecture Search: Hierarchies of Functions with Random Weights as a Deep Prior}
\begin{table}[b]
\centering
\resizebox{0.95  \columnwidth}{!}{%
\begin{tabular}{lccc}
 & MNIST & Fashion & CIFAR10 \\ 
\toprule
Linear Classifier (LC) & \textbf{91.48} & \textbf{85.91} & \textbf{41.12} \\ 
\midrule
Random LeNet + LC  & \textcolor{red}{88.76} & \textcolor{red}{80.33} & \textcolor{blue}{43.40} \\ 
Trained LeNet + LC & 98.73 & 90.89 & 58.92 \\ 
\midrule
Random CNN-2L + LC & \textcolor{blue}{98.01} & \textcolor{blue}{89.29} &  \textcolor{blue}{60.26} \\ 
Trained CNN-2L + LC & 98.86 & 92.13 & 70.86 \\ 
\end{tabular}%
}
\smallskip
\caption{Example classification accuracies (in \%) when computing the randomly projected embedding and training a subsequent linear classifier are compared to training the latter directly on the raw input data (bold reference value). Values are colored in blue if the random weight deep prior facilitates the task. Red values illustrate a disadvantage. Fully trained architecture accuracies are provided for completeness. Example architectures are the three convolutional layer LeNet \cite{LeCun1998} with average pooling and a similar two-layer convolutional architecture with max pooling. \label{tab:prior_intro}}
\end{table}
In the recent work of \cite{Ulyanov2020}, the authors investigate the role of convolution neural networks' structure in image generation and restoration tasks. Their key finding is that a significant amount of image statistics is captured by the structure, even if a network is initialized randomly and subsequently trained on individual image instances, rather than data populations. Quoting the authors "the structure of the network needs to resonate with the structure of the data" \cite{Ulyanov2020}, referred to as a \textit{deep image prior}. We adopt this terminology. 

To start with an intuitive picture behind such a deep prior, we first conduct a small experiment to showcase the effect in our classification context, before referring to previous theoretical findings. Specifically, in table \ref{tab:prior_intro}, we take three popular image classification datasets: MNIST \cite{LeCun1998}, FashionMNIST \cite{Xiao2017} and CIFAR-10 \cite{Krizhevsky2009}, and proceed to train a single linear classification layer to convergence. We report the accuracy on the test set in bold print in the first row and then repeat the experiment in two further variations. \\ 
First, we compute two convolutional neural architectures with random weights, drawn from Gaussian distributions according to \cite{He2015}, before again training a single linear classification layer to convergence on this embedding. One of these architectures is the popular LeNet \cite{LeCun1998} with 3 convolutional layers, intermittent average pooling operations and rectified linear unit (ReLU) activation functions. The other is a 2 layer convolutional layer architecture, again with ReLUs, but with max pooling operations. For convenience we have color coded the results in red and blue, red if the obtained accuracy is worse than simply training the linear classifier on the raw input, blue if the result is improved. We can observe that the 2 layer max-pool architecture with random weights dramatically improves the classification accuracy, even though none of the encoder's weights were trained. \\
In a second repetition of the experiment we also train the convolutional neural architectures' weights to full convergence. Expectedly, this improves the performance. However, we can also observe that the gap to the random encoder version of the experiment is less than perhaps expected. 

Before proceeding to further build on these preliminary results, we highlight two seminal works, which detail the theoretical understanding behind the values presented in table \ref{tab:prior_intro}. 
The work of Saxe \etal \cite{Saxe2011} has proven that the combination of random weight convolutions and pooling operations can have inherent frequency selectivity with well-known local translation invariance. Correspondingly, they conclude that large portions of performance in classification stems from the choice of architecture, similar to observations of \cite{Ulyanov2020} for generation. The work by Giryes \etal \cite{Giryes2016} further proves that deep neural networks with random i.i.d. Gaussian weights preserve the metric structure of the data throughout the layer propagation. For the specific case of piecewise linear activations, such as ReLU, they further show that a sensitivity to angles is amplified by modifying distances of points in successive random embedding. This mechanism is suspected to draw similar concepts closer together and push other classes away, promoting separation. 

\subsection{Deep Prior Neural Architecture Search}
\begin{figure*}
	\centering
	\includegraphics[width=0.99 \textwidth]{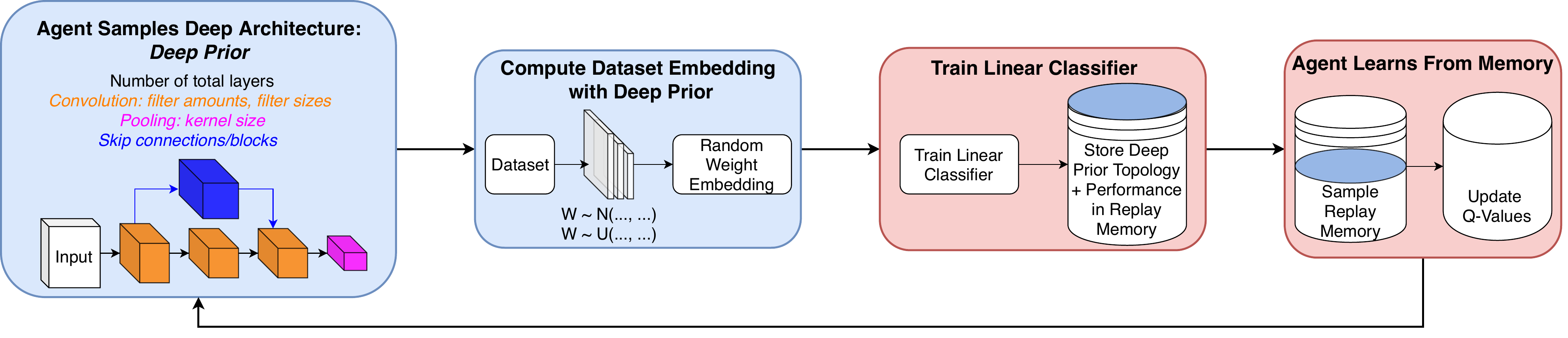}
	\caption{Illustration of the steps involved in Deep Prior Neural Architecture Search (DP-NAS). In DP-NAS, a reinforcement learning q-agent is trained to learn to assemble suitable architectures with random weights across a diverse range of options (illustrated in the first box). In contrast to conventional NAS, these architectures do not have their weights trained. Instead, a randomly projected dataset embedding is calculated (second box) and only a linear classifier is trained (third box). The essential premise is that suitable deep priors contain functional forms of selectivity with respect to low level data statistics, such that classes are easily linearly separable in the transformed space. The q-agent learns to find the latter for a given task (fourth box with outer loop). For convenience, parts of the algorithm that do not involve any parameter inference are colored in blue, with red parts illustrating the only trained components.  \label{fig:DPNAS}}
\end{figure*}

Leveraging previously outlined theoretical findings and encouraged by our initial ablation experiment, we formulate the first central hypothesis of this work:
\begin{tcolorbox}[colback=green!5!white,colframe=green!35!white,title={Hypothesis 1 - deep prior neural architecture search}, coltitle=black]
  A hierarchical neural network encoder with random weights acts as a \textit{deep prior}. We conjecture that there exist deep priors, which we can discover through modern architecture search techniques, that lead to a classification task's solution to the same degree of a fully trained architecture. 
\end{tcolorbox}
A crucial realization in the practical formulation of such a deep prior neural architecture search (DP-NAS) is that we are not actually required to infer the weights of our deep neural architectures. Whereas previous applications of NAS \cite{Baker2016, Real2017, Zoph2017} have yielded impressive results, their practical application remains limited due to the excessive computation involved. This is because neural architecture search, in independence of its exact formulation, requires a reward signal to advance and improve. For instance, the works of Baker \etal \cite{Baker2016} or Zoph \etal \cite{Zoph2017} require full training of each deep neural network to use the validation accuracy to train the agent that samples neural architectures. Reported consumed times for a search over thousands of architectures are thus regularly on the order of multiple weeks with tens to hundreds of GPUs used. Our proposed DP-NAS approach follows the general formulation of NAS, alas significantly profits from the architecture weights' random nature. 

We visualize our procedure in figure \ref{fig:DPNAS}. In essence, we adopt the steps of MetaQNN \cite{Baker2016}, without actual deep neural network training. It can be summarized in a few steps:
\begin{enumerate}
\item We sample a deep neural architecture and initialize it with random weights from Gaussian distributions.
\item We use this deep prior to calculate a single pass of the entire training dataset to compute its embedding.
\item The obtained transformed dataset is then used to evaluate the deep prior's suitability by training a simple linear classifier. Based on a small held-out set of training data, the latter's validation accuracy is stored jointly with the deep prior topology into a replay buffer.
\item The current architecture, together with random previous samples stored in the replay buffer, are then used to update the q-values of a reinforcement learner.
\end{enumerate}
Once the search advances, we progressively decrease an epsilon value, a threshold value for a coin flip that determines whether a deep prior is sampled completely randomly or generated by the trained agent, from unity to zero. 

To get a better overview, we have shaded the parts of figure \ref{fig:DPNAS} that require training in red and parts that do not in blue. As the sampling of a neural architecture is computationally negligible and a single computation of the deep prior embedding for the dataset on a single GPU is on the order of seconds, the majority of the calculation is now redirected to the training of a linear classifier and updating of our q-agent. Fortunately, the former is just a matrix multiplication, the latter is a mere computation of the Bellman equation in tabular q-learning. The training per classifier thus also resides in the seconds regime. Our DP-NAS code is available at: \url{https://github.com/ccc-frankfurt/DP-NAS}

\subsection{Ablation study: DP-NAS on FashionMNIST}\label{sec:hyper-params}
To empirically corroborate hypothesis 1, we conduct a DP-NAS over 2500 architectures on FashionMNIST. Here, the related theoretical works, introduced in the beginning of this section, serve as the main motivation behind our specific search space design. Correspondingly, we have presently limited the choice of activation function to ReLUs and the choice of pooling operation to max pooling. We search over the set of convolutions with $\{ 16, 32, 64, 128, 256, 512, 1024 \}$ random Gaussian filters, drawn according to \cite{He2015}, of size $\{ 1, 3, 5, 7, 9, 11 \}$ with options for strides of one or two. Similarly, potential receptive field sizes and strides for pooling are both $\{ 2, 3, 4 \}$. We allow for the sampling of skip connections in the style of residual networks \cite{He2016, Zagoruyko2016}, where a parallel connection with convolution can be added to skip two layers. We presently search over architectures that contain at least one and a maximum of twelve convolutional layers. The subsequent linear classifier is trained for 30 epochs using an Adam optimizer \cite{Kingma2015} with a learning rate of $10^{-3}$ and a mini-batch size of $128$. We have applied no additional pre-processing, data augmentation, dropout, weight decay regularization, or similar techniques. We start with a 1500 long exploration phase, $\epsilon = 1$, before starting to exploit the learned q-agent by reducing $\epsilon$ by 0.1 every subsequent 100 architectures. 

\begin{figure}
	\includegraphics[width=\columnwidth]{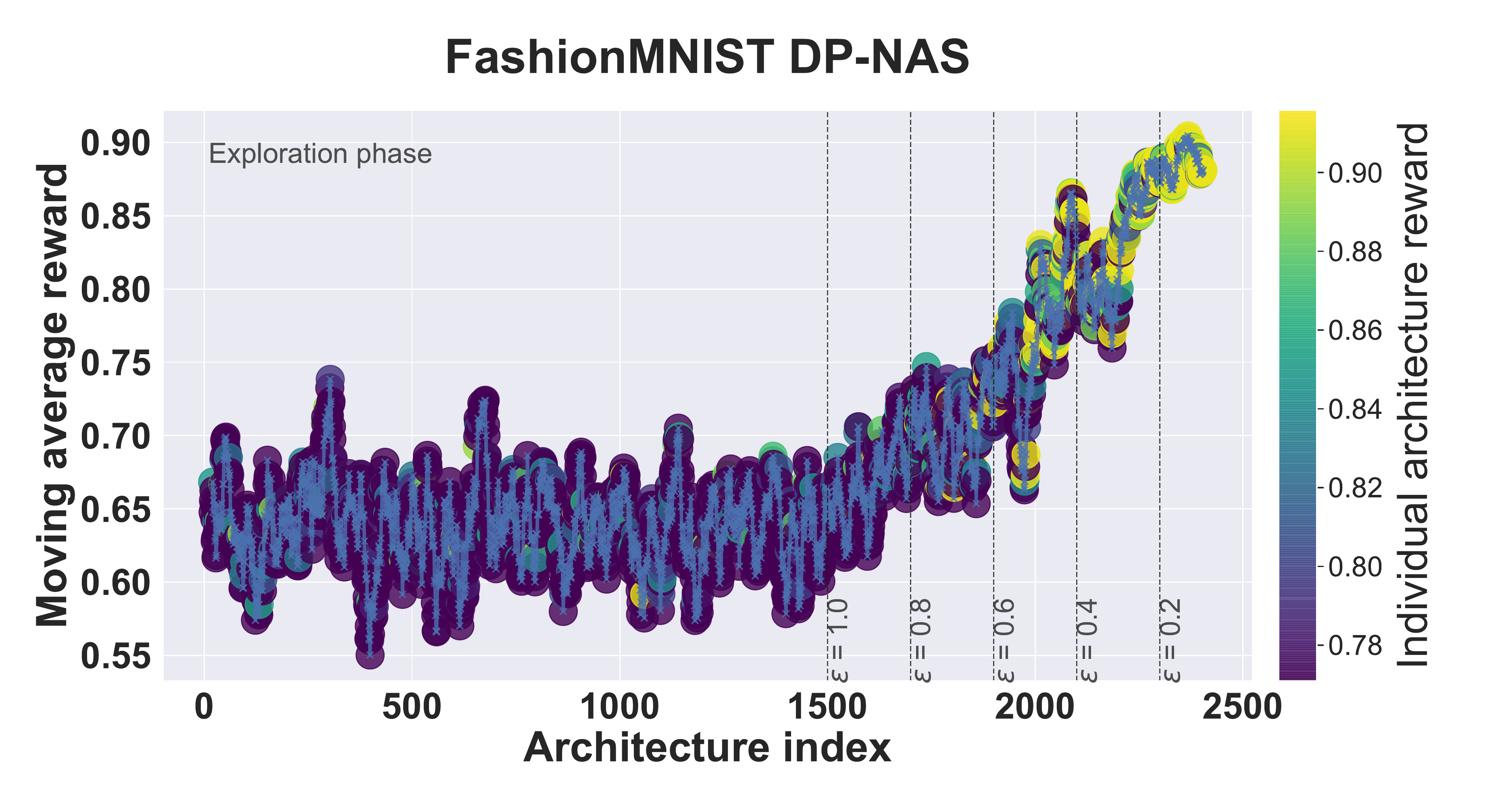}
	\caption{\label{fig:DPNAS_Fashion} DP-NAS on FashionMNIST. The y-axis shows the moving average reward, \ie rolling validation accuracy, whereas the color of each architecture encodes the individual performance. Dashed vertical lines show the used epsilon greedy schedule. }
\end{figure}
Figure \ref{fig:DPNAS_Fashion} shows the obtained DP-NAS results. The graph depicts the mean average reward, \ie the rolling validation accuracy over time, as a function of the number of sampled architectures. Vertically dashed lines indicate the epsilon greedy schedule. In addition, each plotted point is color coded as to represent the precisely obtained accuracy of the individual architecture. From the trajectory, we can observe that the agent successfully learns suitable deep priors over time. The best of these deep priors enable the linear classifier to reach accuracies around $92 \%$, values typically reached by fully trained networks without augmentation. 

At this point, we can empirically suspect these results to already support hypothesis 1. To avoid jumping to premature conclusions, we further corroborate our finding by investigating the role of the particularly drawn random weights from their overall distribution, as well as an experiment to confirm that the best found deep priors do in fact not improve significantly with additional parameter inference. 

\subsection{Did we get lucky on the initialization lottery?}
We expand our experiment with a further investigation with respect to the role of the precisely sampled weights. In particular, we wish to empirically verify that our found deep priors are in fact deep priors. In other words, the observed success is actually due to the chosen hierarchy of functions projecting into spaces that significantly facilitate classification, rather than being subject to lucky draws of good sets of precise weight values when sampling from a Normal distribution. Our second hypothesis is thus:
\begin{tcolorbox}[colback=green!5!white,colframe=green!35!white,title={Hypothesis 2 - deep priors and initialization lottery}, coltitle=black]
  A hierarchical neural network encoder with random weights acts as a \textit{deep prior}, irrespectively of the precisely sampled weights. In particular, it is not subject to an initialization lottery. 
\end{tcolorbox}
This train of thought is motivated from recent findings on the lottery ticket hypothesis \cite{Frankle2019}, where it is conjectured that there exists an initialization lottery in dense randomly initialized feed-forward deep neural networks. According to the original authors, winning this lottery is synonymous with finding an initial set of weights that enables training.

\begin{figure}
	\includegraphics[width=\columnwidth]{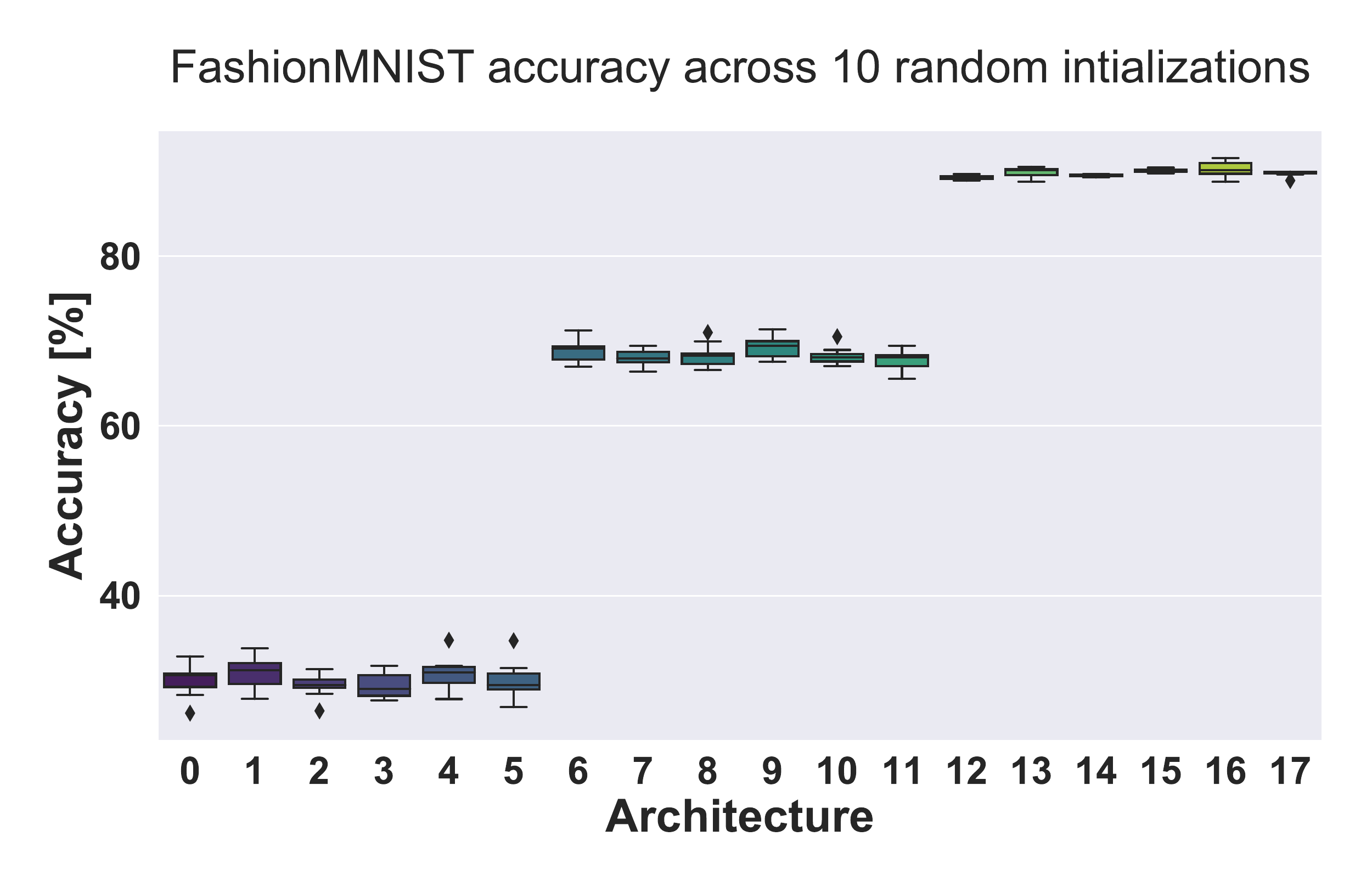}
	\caption{\label{fig:weight_reinit} Accuracy for 18 deep priors  across 10 experimental repetitions with randomly sampled weights. Six deep priors for respective 3 performance segments of figure \ref{fig:DPNAS_Fashion} (0-5 low, 6-11 median, 12-17 top) have been selected. Result stability suggests that the architecture composition is the primary factor in performance. }
\end{figure}

To empirically confirm that the structure is the imperative element, we randomly select 18 deep priors from our previous FashionMNIST search. 6 of these are sampled from the lowest performing architectures, 6 are picked around the median, and 6 are chosen to represent the top deep priors. For each of these deep priors, we repeat the linear classifier training for 10 independently sampled sets of weights. The respective figure \ref{fig:weight_reinit} shows the median, upper and lower quartiles, and deviations of the measured test accuracy. We observe that the fluctuation is limited, the results reproducible and the ordering of the deep priors is thus preserved. Whereas minor fluctuations for particular weight samples seem to exist, the results suggest that the architecture itself is the main contributor to obtained performance. 

\subsection{Are Deep Priors predictive of performance with parameter inference?}
\begin{figure}
	\includegraphics[width=\columnwidth]{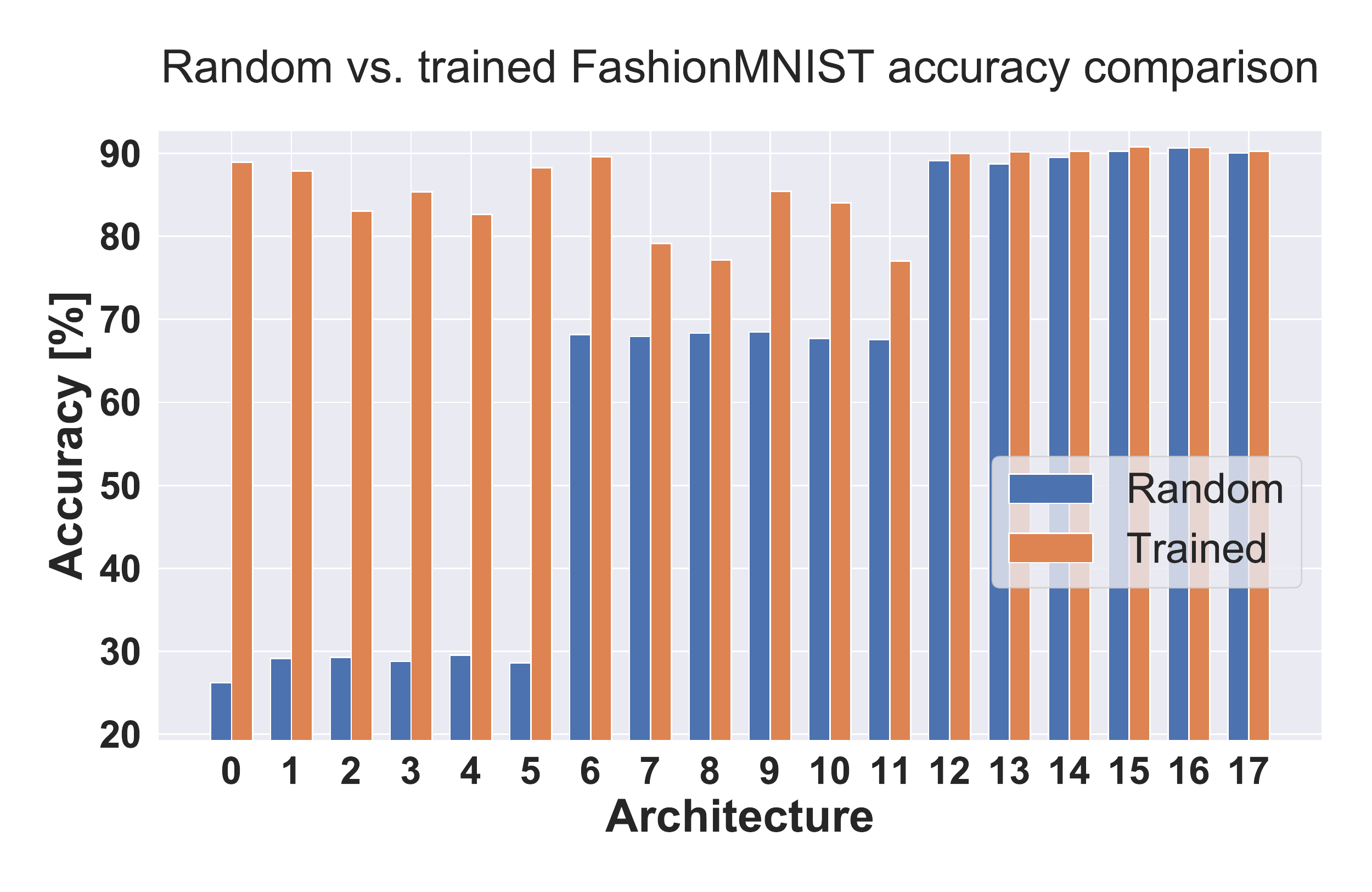}
	\caption{\label{fig:random_vs_trained} Accuracy comparison when only a linear classifier is trained on top of the deep prior and when the entire architecture is trained. The same 18 deep priors as picked in figure \ref{fig:weight_reinit}, representing respective 3 performance segments of figure \ref{fig:DPNAS_Fashion} (0-5 low, 6-11 median, 12-17 top),  are shown. Results demonstrate that the best deep priors enable a solution without parametric inference and performance improvement is negligible with additional training.}
\end{figure}
To finalize our initial set of ablation experiments, we empirically confirm that the best deep priors do "fully resonate" with underlying dataset statistics. If we conjecture the best deep priors to extract all necessary low level image statistics to allow for linear decision boundaries to effectively provide a solution, then we would expect no additional parameter inference to yield significant improvement.

\begin{tcolorbox}[colback=green!5!white,colframe=green!35!white,title={Hypothesis 3 - deep priors and parameter inference}, coltitle=black]
  A deep image prior performs at least equivalently, if not significantly better, when parameters are additionally inferred from the data population. However, we posit that the best deep priors are already close to the achievable performance. 
\end{tcolorbox}

To investigate the above hypothesis, we pick the same 18 architectures as analyzed in the last subsection to again represent the different deep prior performance segments. In contrast to the random weight deep prior experiments, we now also fully train the deep architecture jointly with the classifier. We show the obtained test accuracies in comparison with the deep prior with exclusive training of the linear classifier in figure \ref{fig:random_vs_trained}. We first observe that all trained architectures perform at least as well as their random counterparts. As expected, the full training did not make it worse. Many of the trained architectures with non-ideal deep priors still perform worse with respect to the best untrained deep priors. The latter all improve only very marginally, suggesting that adjustments to the parameters provide negligible value. The best deep priors seem to solve the task to the same degree that a fully trained deep network does. We thus see our third hypothesis to be empirically confirmed, and in turn the initial first hypothesis validated. \\
Although not crucial to our main hypotheses, interestingly, we can also observe that the ordering for the worst to intermediate deep priors in terms of achieved accuracy is not retained. We speculate that this is a consequence of heavy over-parametrization of many constructed deep architectures. The latter entails a high fitting capacity when trained and thus a higher ability to compensate misalignment. 

\section{Continual Learning with Deep Priors}
With the foregoing section focusing on the general feasibility of deep priors, we now extend our investigation towards implications for continual learning. In neural networks the latter is particularly difficult, given that training overwrites parameters towards the presently observed data sub-population: the phenomenon of \textit{catastrophic interference} \cite{McCloskey1989, Ratcliff1990}. As highlighted in recent reviews \cite{Parisi2019, Mundt2020}, the challenge is already significant when considering simple \textit{class incremental} tasks, such as the prevalent scenario of splitting datasets into sequentially presented disjoint sets of classes. In contrast, we formulate our central hypothesis with respect to continual learning with deep priors:

\begin{tcolorbox}[colback=green!5!white,colframe=green!35!white,title={Hypothesis 4 - deep priors and continual learning}, coltitle=black]
  If classes in a dataset originate from the same data distribution, finding a deep prior on a subset of dataset classes can be sufficient to span prospective application to the remainder of the unseen classes.
\end{tcolorbox}

The above hypothesis is motivated from the expectation that a deep prior primarily captures the structure of the data through low-level image statistics. The latter can then safely be considered to be shared within the same data distribution. In particular, the hypothesis seems reasonable with hindsight knowledge of the findings in prior theoretical works, which we re-iterate to have proven that Gaussian random projections enhance seprability through inherent frequency and angular sensitivity \cite{Candes2006,Saxe2011,Giryes2016}. A deep prior, found to respond well to a subset of classes, can then also be assumed to act as a good prior for other labels of instances drawn from the same data distribution. To give a practical example, we posit that a deep prior found for the t-shirt and coat classes transfers to images of pullovers and shirts under shared sensor and acquisition characteristics. 

From this hypothesis, an intriguing consequence arises for deep continual learning. For instance, learning multiple disjoint tasks in sequence, that is sharing the neural network feature extractor backbone but training a separate task classifier, struggles with catastrophic interference in conventional training because the end-to-end functional is subject to constant change. In contrast, if the deep prior holds across these tasks, the solution is of trivial nature. Due to the absence of tuning the randomly initialized weights, by definition, the deep prior encoder is not subject to catastrophic interference in such continual data scenarios. 
We start with an empirical investigation of the practical validity of the above hypothesis in this scenario. With empirical confirmation in place, we then proceed to extend this investigation to a more realistic scenario, where a single task of continuously growing complexity is considered. We posit that a deep prior significantly facilitates this more generic formulation, as we only need to regulate inference at the prediction level.  

\subsection{Preliminaries: scenarios and methods}
Before delving into specific experiments, we provide a short primer on predominantly considered assumptions and result interpretation, in order to place our work in context.
We do not at this point provide a review on the vast continual learning literature and defer to the referenced surveys. \\

\noindent \textbf{Continual learning scenarios} \\
For our purposes of continual learning with deep priors, we investigate two configurations:
\begin{enumerate}
\item \textit{Multi-head incremental learning:} in this commonly considered simplified scenario, sets of disjoint classes arrive in sequence. Here, each disjoint set presents its own task. Assuming the knowledge of a task id, separate prediction layers are trained while attempting to preserve the joint encoder representations for all tasks. Although often considered unrealistic, complex strategies are already required to address this scenario.
\item \textit{Single-head incremental learning:} the scenario mirrors the above, alas lifts the assumption on the presence of a task identifier. Instead of inferring separate predictors per task, a single task is extended. The prediction layer's amount of classes is expanded with every arrival of a new disjoint class set. In addition to catastrophic interference in a deep encoder, interference between tasks is now a further nuisance. For instance, a softmax based linear prediction will now also tamper with output confidences of former tasks.
\end{enumerate}

We do not add to ongoing discussions on when specific scenario assumptions have practical leverage, see \eg \cite{Farquhar2018a} for the latter. In the context of this work, the multi-head scenario is compelling because a separate classifier per task allows to directly investigate hypothesis four. Specifically, we can gauge whether deep priors found on the first task are suitable for prospective tasks from the same distribution. For the single-head scenario, we need to continuously train a linear prediction layer on top of the deep prior. As such, we will also need to apply measures to alleviate catastrophic interference on this single learned operation. However, in contrast to maintaining an entire deep encoder, we would expect this to work much more efficiently. \\

\noindent \textbf{A brief primer on interpreting reported results} \\
Independently of the considered scenario, techniques to address catastrophic interference typically follow one of three principles: explicit parameter regularization \cite{Kirkpatrick2017, Lee2017, Chaudhry2018}, retention and rehearsal of a subset of real data (a core set) \cite{Rebuffi2017, Nguyen2018}, or the training of additional generative models to replay auxiliary data \cite{Ayub2021, Mundt2019, Ostapenko2019, Shin2017}. For the latter two families of methods, the stored or generated instances of older tasks get interleaved with new task real data during continued training. Once more, we defer to the survey works for detailed explanations of specific techniques \cite{Parisi2019, Mundt2020}. For our purposes of demonstrating an alternative to researching continual deep learning from a perspective of catastrophic interference in deep encoders, it suffices to know that all these methods train encoders and construct complex mechanisms to preserve its learned representations. 

We point out that contrasting performances between these techniques in trained deep neural networks is fairly similar to a comparison of apples to oranges. Whereas the essential goal is shared, the amount of used computation, storage or accessibility of data varies dramatically. Correspondingly, \textit{in our result tables and figures we provide a citation to each respective technique and an additional citation next to a particular accuracy value to the work that has reported the technique in the specific investigated scenario}. 
Whereas we report a variety of these literature values, we emphasize that our deep prior architectures do not actually undergo any training. Our upcoming deep prior results should thus be seen from a perspective of providing an alternate way of thinking about the currently scrutinized continual learning challenges. In fact, we will observe that our deep prior experiments yield remarkably competitive performances to sophisticated algorithms. 

\subsection{Disjoint tasks: deep priors as a trivial solution}
\begin{table}
\centering
\begin{tabular}{lcc}
 & \multicolumn{2}{c}{Multi-head Accuracy [\%]} \\ 
 \cmidrule{2-3}
Method & \textbf{MNIST} & \textbf{FashionMNIST} \\
\toprule
EWC \cite{Kirkpatrick2017} & 99.3 \cite{Chaudhry2018} & 95.3 \cite{Farquhar2018a} \\
RWalk \cite{Chaudhry2018} & 99.3 \cite{Chaudhry2018} & - \\
VCL + Core \cite{Nguyen2018} & 98.6 \cite{Farquhar2018a} & 97.1 \cite{Farquhar2018a} \\
VCL \cite{Nguyen2018} & 97.0 \cite{Farquhar2018a} & 80.6 \cite{Farquhar2018a} \\
VGR \cite{Farquhar2018a} & 99.3 \cite{Farquhar2018a} & 99.2 \cite{Farquhar2018a}\\
\midrule
DP & 99.79 & 99.37 \\
\end{tabular} 
\smallskip
\caption{Average accuracy across 5 disjoint tasks with 2 classes for FashionMNIST and MNIST. Our deep prior approach provides competitive performance, even though the corresponding DP-NAS has only been conducted on the initial task.  \label{tab:dp_multi_head}}
\end{table} 
We start by investigating hypothesis four within the multi-head incremental classification framework. For this purpose, we consider the disjoint FashionMNIST and MNIST scenarios, where each subsequent independent task is concerned with classification of two consecutive digits. We repeat our DP-NAS procedure once for each dataset \textit{on only the initial task} and thus only the first two classes. We do not show the precise search trajectories as they look remarkably similar to figure \ref{fig:DPNAS_Fashion}, with the main difference being a shift in accuracy towards a final reached $99.9 \%$ as a result of narrowing down the problem to two classes. Thereafter, we use the top deep prior and proceed to continuously learn linear classification predictions for the remaining classes.

We report the final averaged accuracy across all five tasks in table \ref{tab:dp_multi_head} and compare them with achieved accuracies in prominent literature. We can observe that for these investigated datasets our deep prior hypothesis four holds. With the average final accuracy surpassing 99 \% on both datasets, the originally obtained deep prior for the first task seems to transfer seamlessly to tasks two to five. By definition, as predictions of disjoint tasks do not interfere with each other, a trivial solution to this multi-head continual learning scenario has thus been obtained. This is in stark contrast to the referenced complexly tailored literature approaches. 

\subsection{Alleviating catastrophic interference in a single prediction layer on the basis of a deep prior}
\begin{figure}
	\includegraphics[width=\columnwidth]{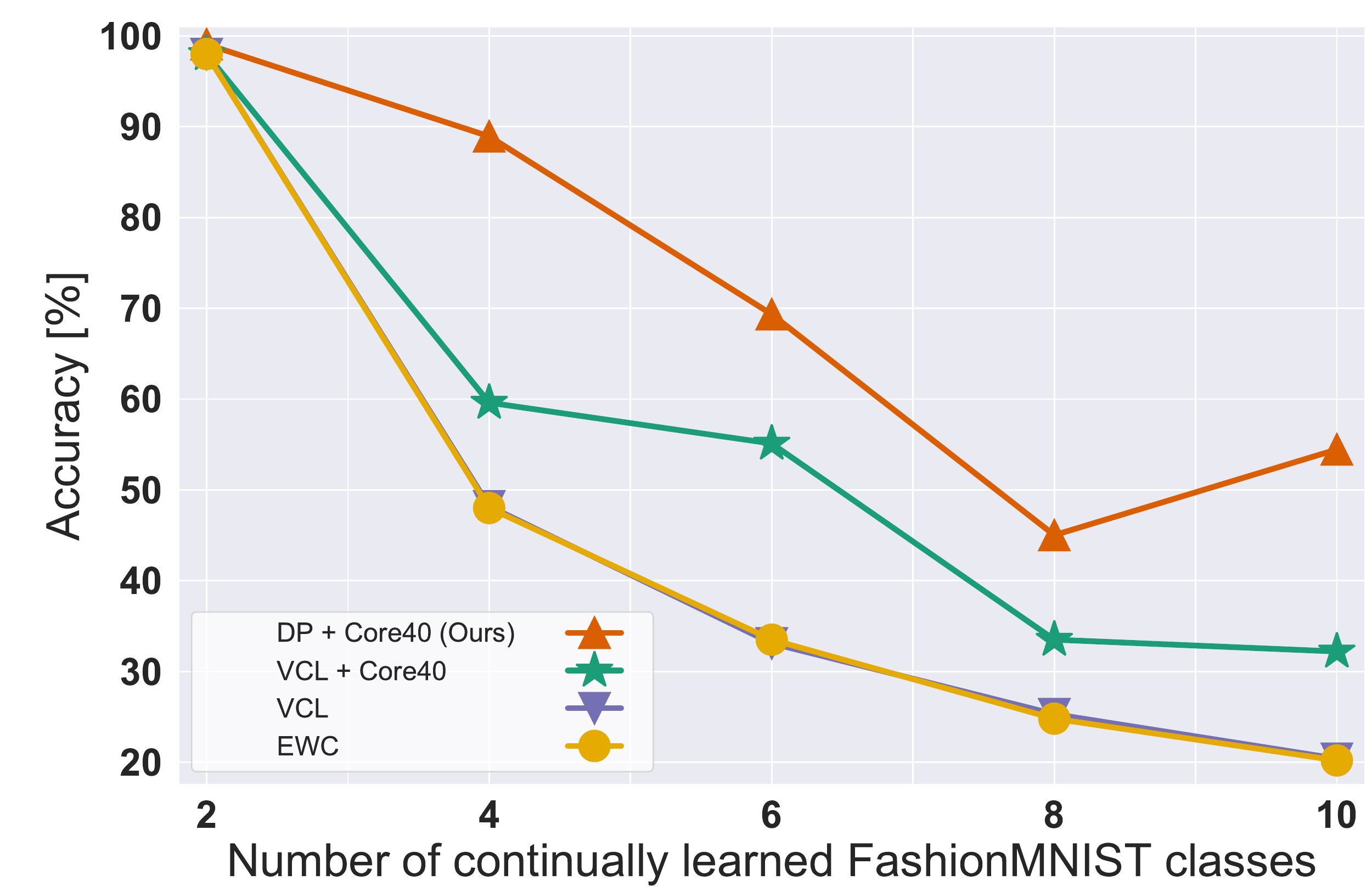} 
	\caption{Single-head FashionMNIST accuracy. \label{fig:Fashion_CL}}
\end{figure}
To further empirically corroborate our conjecture of hypothesis four, we conduct additional class incremental continual learning experiments in the single-head scenario. This scenario softens the requirement of task labels by continuing to train a single joint prediction layer that learns to accommodate new classes as they arrive over time. Whereas our deep prior (again only obtained on the initial two classes) does not require any training, we thus need to limit catastrophic interference in our linear classifier. Note that we can use any and all of the existing continual learning techniques for this purpose. However, we have decided to use one of the easiest conceivable techniques in the spirit of variational continual learning (VCL) \cite{Nguyen2018} and iCarl \cite{Rebuffi2017}. The authors suggest to store a core set, a small sub-set of original data instances, and continuously interleave this set in the training procedure. Although they suggest to use involved extraction techniques, such as k-center or herding, we sample uniformly in favor of maximum simplicity. In contrast to prior works, as the deep prior remains untrained, we have the option to conserve memory by storing randomly projected embeddings instead of raw data inputs. 

\subsubsection{FashionMNIST revisited: single-headed}
\begin{figure}
	\includegraphics[width=\columnwidth]{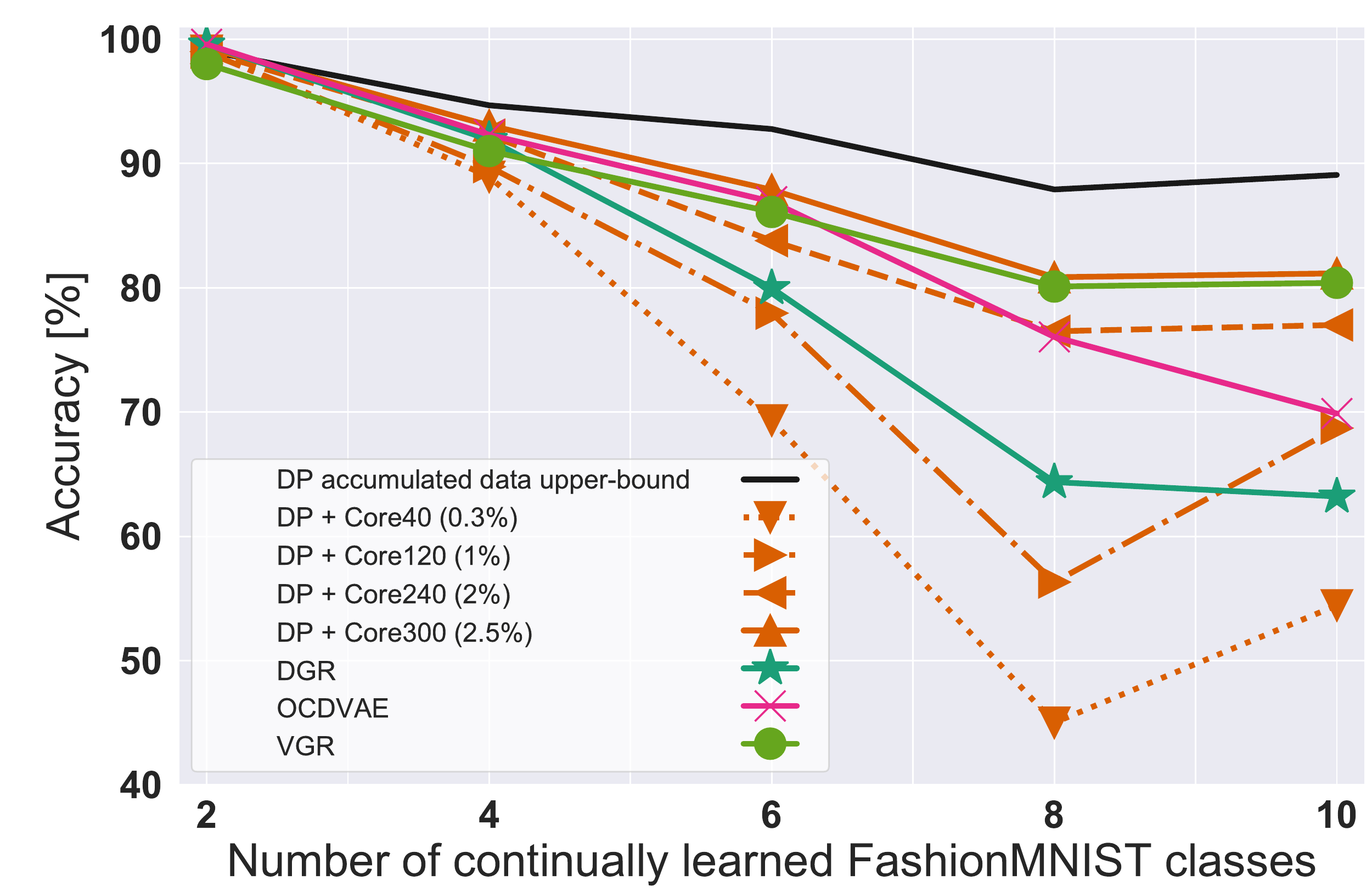} 
	\caption{Single-head FashionMNIST for varying core set sizes. \label{fig:Fashion_CL_Core}}
\end{figure}
Our first set of single-head incremental classification experiments uses the deep prior found for FashionMNIST on the initial two classes, similar to our previous multi-head experimental section. To protect the single linear prediction, we store a random core set of 40 examples, in correspondence with prior experiments in the literature \cite{Farquhar2018a}. Figure \ref{fig:Fashion_CL} provides the respective result comparison. Among the reported techniques, storing 40 exemplars in a deep prior seems to significantly outperform storing 40 exemplars in VCL. Naturally, this is because VCL requires protection of all representations in the entire neural network, whereas the deep prior is agnostic to catastrophic interference and only the single layer classifier needs to be maintained. The popular parameter regularization technique Elastic Weight Consolidation (EWC) \cite{Kirkpatrick2017} is similarly outperformed. 

We nevertheless observe a significant amount of performance degradation. To also corroborate our hypothesis four in this single-head scenario and show that this forgetting can be attributed exclusively to the linear classifier, we further repeat this experiment with increased amount of instances stored in the core set. For ease of readability, these additional results are shown in figure \ref{fig:Fashion_CL_Core} for a stored amount of 40, 120, 240, 300 instances, corresponding to a respective $0.33 \%, 1 \%, 2 \%$ and $2.5 \%$ of the overall data. The black solid curve shows the achieved accuracy when all real data is simply accumulated progressively. Our first observation is that the final accuracy of the latter curve is very close to the final performance values reported in the full DP-NAS of figure \ref{fig:DPNAS_Fashion}, even though we have only found a deep prior on the first two classes. Once more, we find additional evidence in support of hypothesis four. This is further substantiated when observing the curves for the individual core set sizes. We can see that in the experiments with 2 and 2.5 \% stored data instances, our deep prior beats very complex generative replay techniques. All three reported techniques: deep generative replay (DGR) \cite{Shin2017}, variational generative replay (VGR) \cite{Farquhar2018a} and open-set denoising variational auto-encoder (OCDVAE) \cite{Mundt2019} employ additional deep generative adversarial networks or variational auto-encoders to generate a full sized dataset to train on. Although this is an intriguing topic to explore in the context of generative modelling, our experiments indicate that for classification purposes our simple deep prior approach seems to have the upper hand. 

\subsubsection{The easier MNIST and more difficult CIFAR-10}
\begin{table}
\centering
\begin{tabular}{lccc}
 Accuracy [\%]  & \textbf{MNIST}  & \multicolumn{2}{c}{\textbf{CIFAR-10}} \\ 
 \cmidrule{2-4}
Method & A10, D2 & A5, D1 & A10, D5 \\
\toprule
EWC \cite{Kirkpatrick2017} & 55.80 \cite{Chaudhry2018} & - & 37.75 \cite{Hu2019} \\
IMM \cite{Lee2017} & 67.25  \cite{Hu2019} & 32.36 \cite{Hu2019} & 62.98 \cite{Hu2019}\\
DGR \cite{Shin2017} & 75.47 \cite{Hu2019} & 31.09 \cite{Hu2019} & 65.11 \cite{Hu2019} \\ 
PGMA \cite{Hu2019} & 81.70 \cite{Hu2019} & 40.47 \cite{Hu2019} & 69.51 \cite{Hu2019} \\ 
RWalk \cite{Chaudhry2018} & 82.50 \cite{Chaudhry2018} & - & - \\
iCarl \cite{Rebuffi2017} & 55.80 \cite{Chaudhry2018}  & 57.30 \cite{Ayub2021} & - \\ 
DGM \cite{Ostapenko2019} & - & 64.94  \cite{Ayub2021} & - \\ 
EEC  \cite{Ayub2021} & - & 85.12 \cite{Ayub2021} & - \\ 

\midrule
DP + core & 76.31 & 58.13 & 65.15 \\
\end{tabular} 
\smallskip
\caption{Average final single-head accuracy. MNIST is reported on 10 classes, after 5 increments containing two classes (A10, D2). For CIFAR-10, accuracies on 5 classes with class increments of size 1 (A5, D1), and on all 10 classes after two increments of 5 classes (A10, D5) are shown. Used core set sizes are 10 instances per task for MNIST, following the experiments of \cite{Chaudhry2018}, and a total memory of 2000 instances for CIFAR-10, according to iCarl \cite{Rebuffi2017}. \label{tab:dp_single_head}}
\end{table}
We finalize our experiments with an additional investigation of the MNIST and CIFAR-10 datasets in the single-head scenario. Once more, we follow previous literature and store 10 randomly sampled data instances per task for MNIST, as suggested in Riemannian walk \cite{Chaudhry2018}, and a maximum memory buffer of size 2000 for CIFAR-10, as suggested in iCarl \cite{Rebuffi2017}. Following our hypothesis, the DP-NAS is again conducted exclusively on the initial classes. 

In table \ref{tab:dp_single_head} we report the final obtained accuracy and compare it to literature approaches. Similar to VCL with core sets, we observe that storing a core set for our linear classifier on top of the deep prior outperforms the core set approach in iCarl. In comparison with the remaining methods we can see that the simple deep prior approach surpasses multiple methods and only falls behind a few select works. The latter can be attributed to additional storage and auxiliary model assumption. For instance, in EEC \cite{Ayub2021} an additional generative model learns to replay the full dataset embeddings. This procedure could find straightforward transfer to our randomly projected deep prior embeddings and is left for future work. For now, we conclude our work by highlighting that there exist alternate methods without parameter inference as potential solutions to explore for both deep neural network classification and continual learning. 

\section{Limitations and prospects}
\noindent \textbf{Domain incremental scenarios:} Above continual learning experiments are limited in that they do not consider sequential data stream scenarios where arriving data $x$ no longer is drawn from the same data distribution, \ie the domain $p_{t}(x) \neq p_{t+1}(x)$ \cite{Delange2021}. Previously postulated hypothesis four can no longer be expected to hold due to potential changes in image statistics. However, one could define a progressive version of DP-NAS, such that the random architecture found for the initial distribution is extended with functions and connections to accommodate further distributions. \\

\noindent \textbf{Search space transformations}: In similar spirit to the aforementioned point, other datasets may require more than the current angular and frequency selectivity of convolutional ReLU blocks. An intriguing future direction will be to explore random deep neural networks in a wider sense, with flexible activation functions or even interpretable transformations added to the search space. This could in turn provide a chance for a more grounded understanding of the relationship between the data distribution and the necessary transformations to accomplish a certain task. \\

\noindent \textbf{Fully catastrophic interference free classification:} For our current single-head continual learning experiments we have optimized a growing linear softmax classifier. Naturally this single prediction layer still suffers from catastrophic interference that needs to be alleviated. It will be interesting to lift this by examining generative alternatives, \eg distance based decisions or mixture models.   \\

\noindent \textbf{Deterministic vs. stochastic deep prior:} Our current deep priors do not fully leverage the weight distributions. After weights are sampled, the deep prior is treated as a deterministic processing block. As we do not train the deep prior, we conjecture that full sampling of weights with propagation of uncertainties, in the sense of Bayesian neural networks \cite{Blundell2015}, can provide additional crucial information. \\

\noindent \textbf{Autoencoding, inversion and compression:} The present focus has been on classification. However, there also exists prior work on weight-tied random autoencoders \cite{Li2019}. \cite{Shu2017} state that random autoencoders work surprisingly well, due to the symmetric and invertible structure, discarding only information on color, but preserving the one on geometry. In a similar vein to the experiments conducted in this paper, an appropriate deep prior could thus also be searched for.

\section{Conclusion}
In this paper we have analyzed the classification performance of the neural network structure independently of parametric inference. Using the proposed deep prior neural architecture search, we have shown that it is possible to find random weight architectures that rival their trained counterparts. Further experiments in continual learning lay open promising novel research directions that pursue an entirely different path from the present focus on catastrophic interference in deep encoder architectures. 

\clearpage

{\small
\bibliographystyle{ieee_fullname}
\bibliography{references}
}

\end{document}